\documentclass[10pt]{igs}

\usepackage{tabularray}
\usepackage{multirow}
\usepackage{graphicx}
\usepackage{comment}
\usepackage{natbib}
\usepackage{xcolor}
\usepackage{graphics}
\usepackage{graphicx}
\usepackage[font=bf]{caption}
\usepackage{hhline}
\usepackage{bigstrut}
\usepackage{graphicx,subcaption,ragged2e}
\usepackage{colortbl}
\usepackage[utf8]{inputenc}
\usepackage{float}
\usepackage{subcaption}
\usepackage{setspace}
\usepackage{adjustbox}
\usepackage{amsmath,amstext,amsopn,amssymb,amsbsy}

\begin{document}

\begin{frontmatter}
%
\begin{center}
\title{Exploring the Potential of Robot-Collected Data for Training Gesture Classification Systems}


%
%
\author{Alejandro GARCIA-SOSA}
\author{, Jose J. QUINTANA}
\author{, Miguel A. FERRER}
\author{and Cristina CARMONA-DUARTE}

\address{Instituto para el Desarrollo Tecnológico y la Innovación en Comunicaciones IDeTIC\\ 
Universidad de Las Palmas de Gran Canaria, Spain\\
\{alejandro.sosa, josejuan.quintana, miguelangel.ferrer, cristina.carmona\}@ulpgc.es
}
\end{center}

%
\maketitle              

\end{frontmatter}

\begin{abstract}

Sensors and Artificial Intelligence (AI) have revolutionized the analysis of human  movement, but the scarcity of specific samples presents a significant challenge in training intelligent systems, particularly in the context of diagnosing neurodegenerative diseases. This study investigates the feasibility of utilizing robot-collected data to train classification systems traditionally trained with human-collected data. As a proof of concept, we recorded a database of numeric characters using an ABB robotic arm and an Apple Watch. We compare the classification performance of the trained systems using both human-recorded and robot-recorded data. Our primary objective is to determine the potential for accurate identification of human numeric characters wearing a smartwatch using robotic movement as training data. The findings of this study offer valuable insights into the feasibility of using robot-collected data for training classification systems. This research holds broad implications across various domains that require reliable identification, particularly in scenarios where access to human-specific data is limited.

\end{abstract}

\section{Introduction}
Nowadays, sensors of different kinds are now widespread throughout every aspect of our lives.Among others, we can find an accelerometer o inertial measurement units (IMUs) in our mobile devices (phones, smartwatches, and nearly every portable electronic gadget).

These IMUs provide considerable information about the user, enabling a wide range of applications. One example is their use as Human-Machine Interfaces (HMI), allowing actions such as answering a phone call by merely moving the phone close to the ear or drawing in the air using gestures.

In the medical field, several studies utilize IMUs for detecting neurodegenerative diseases (\cite{Neurorehab}), aiding in the rehabilitation process after a stroke (\cite{Stroke}), or as Computer-Aided Diagnosis. These studies rely on Artificial Intelligence (AI) classifiers, such as Neural Networks or Support Vector Machines.

Both using IMUs as HMI and for Computer-Aided Diagnosis requires training to recognize specific patterns in order to correctly identify human gestures in real environments, not just in small isolated tests. However, obtaining data for such situations from humans can be highly specific and challenging.

Using robots to address the scarcity of data may prove useful, as robots can be programmed to mimic human movements. Not only can they imitate the movements of healthy individuals, but with the correct parameterization for specific conditions, we can replicate movements seen in individuals with Parkinson's disease, ADHD, and other conditions. This approach could help create substantial databases for training AI systems, with tens or hundreds of thousands of samples, without the need to search for humans with specific characteristics. To achieve this, it will be necessary to synthesize and adjust the robot accurately to replicate human-like movements as closely as possible.
Previous studies regarding classification using IMU-recorded data, some of them about numeric character classification (\cite{AccelPen, HandGestureRecognition}), have a limitation in the database. 

To our knowledge, no other studies about the classification of human movement using robotic movements for training have been done previously. There are  other studies using non-human data for training intelligence-based systems using Virtual Reality (\cite{Bui202232625}).

Our aim in this paper is to explore whether is possible to train gesture classification systems meant to classify human gestures using only robot gestures. In order to do that, we are recording a database of numeric characters using both a human and a robotic arm, and we are training a classifier using that database.

\section{Methodology}
\subsection{Database}

To train the classifier and determine if it is feasible to train it with robotic data, we require data from both robots and humans for training and validation purposes.

To record the database, an Apple Watch Series 6, and the application WData, developed by our team was used. The data is recorded with a sampling rate of 100 Hz. WData is an app available for testing on TestFlight that allows data to be captured from the watch's IMU at a sampling rate of 100 Hz. It makes use of the CMMotionManager tool, which allows the app to extract acceleration, device altitude, rotation speed, calibrated magnetic fields, and gravity direction from the recorded data. The data collected by the watch is sent to a mobile phone, which then uploads it to iCloud, where it can be accessed from a laptop or PC.

\begin{figure}[h!] 
\centering
\begin{subfigure}[t]{0.25\textwidth}
    \includegraphics[width=\textwidth]{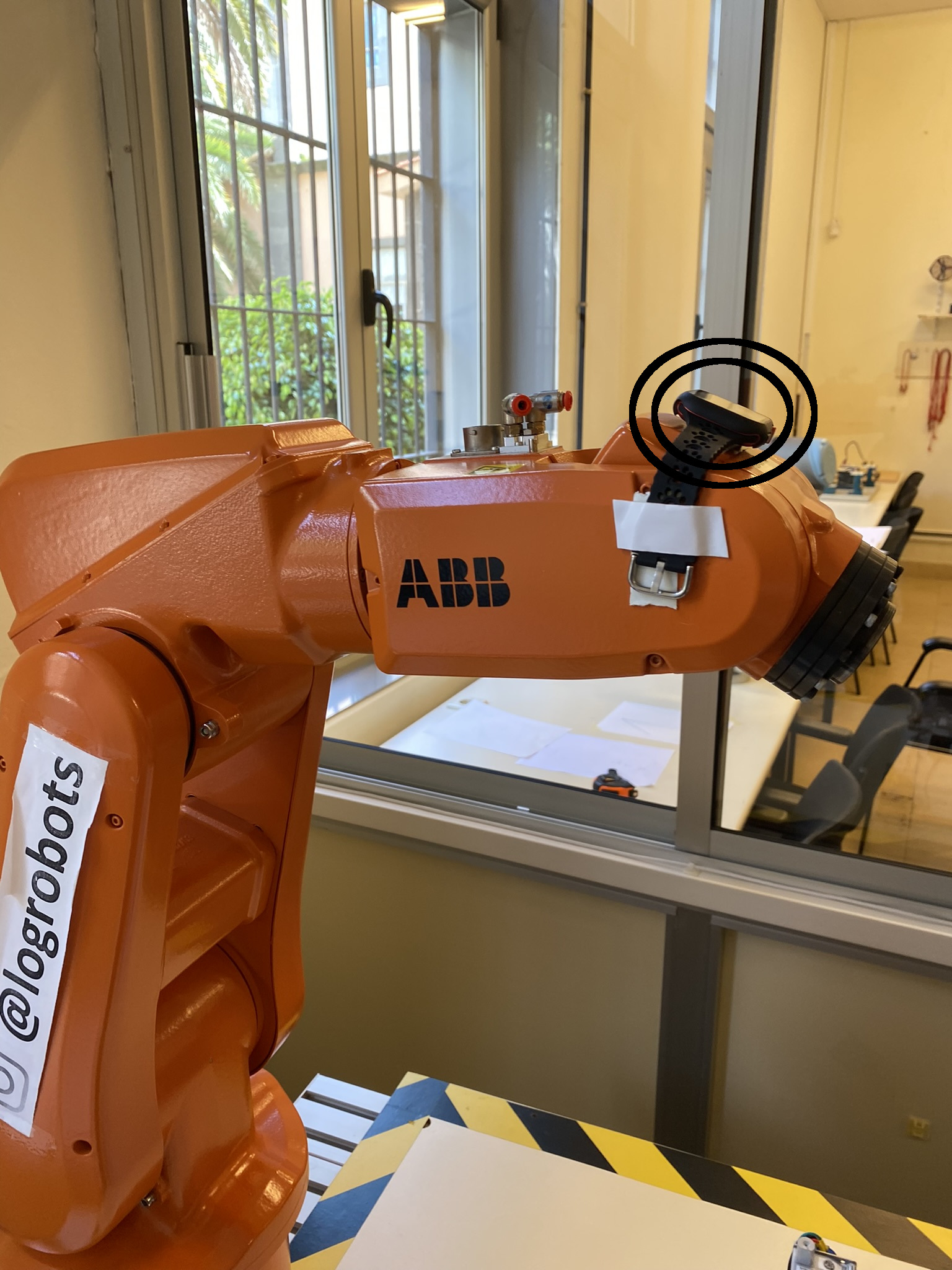}
    \caption{}
\end{subfigure}
\begin{subfigure}[t]{0.25\textwidth}
    \includegraphics[width=\linewidth]{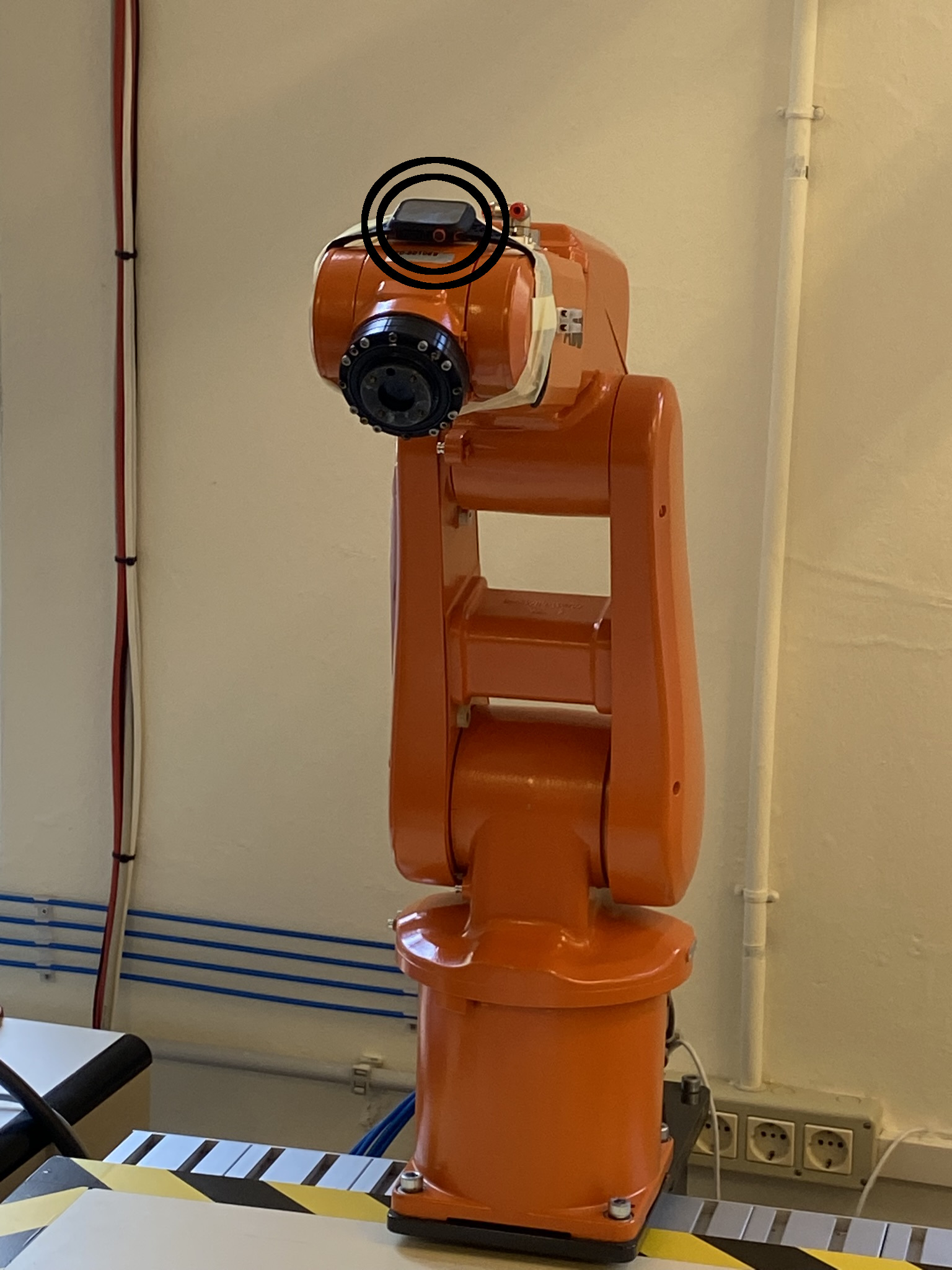}
    \caption{}
\end{subfigure}

\caption{Robot used in the experiments with the smartwatch. }
\label{fig:Robotfoto}
\end{figure}

The human data consists of 100 samples, with 10 samples for each number from 0 to 9. On the other hand, for the robotic movement database, 81 samples of each number were generated and recorded using an ABB IRB120 (see Figure \ref{fig:Robotfoto}). Each sample has a different combination of velocity, wrist angle, size, and rotation, with three variations for each parameter. Both human and robot data were captured with the watch IMU sensors.

To simulate human movement with the robot, it has to be programmed. To this end, human samples were captured by drawing each number ten times, at different velocities and sizes, using wire-type distance sensors. The velocitiy profiles for each number were visually compared, and the ones with less noise for each number were chosen. The recorded data consisted of the coordinates in the three Cartesian axes with a sampling rate of 200Hz (\cite{Quintana2022}). Once the trajectory data was obtained, the procedure followed for the robot to reproduce each combination is as follows: First, the sampling frequency was reduced from 200Hz to 42Hz. Then, for each of the points (x, y, z) of the trajectory, its inverse kinematics was calculated to obtain the equivalent position of each of the robot's joints. Finally, the temporal sequences of joint positions were transferred to the robot, which executed them while the watch captured the data.


After a visual comparison of the human data and the robotic data, some differences were detected in the rotation of the numbers. Since the original robot wrist was set to be parallel to the floor, it needed to be adjusted. After conducting several tests with different rotations, the robotic dataset was re-recorded, incorporating a 60º rotation in the Y-axis, a 5º rotation in the Z-axis, and a 20º rotation in the X-axis. These rotations were intended to represent the natural human wrist position.

\begin{figure}
    \centering
    \begin{subfigure}[t]{0.4\textwidth}
        \includegraphics[width=\textwidth]{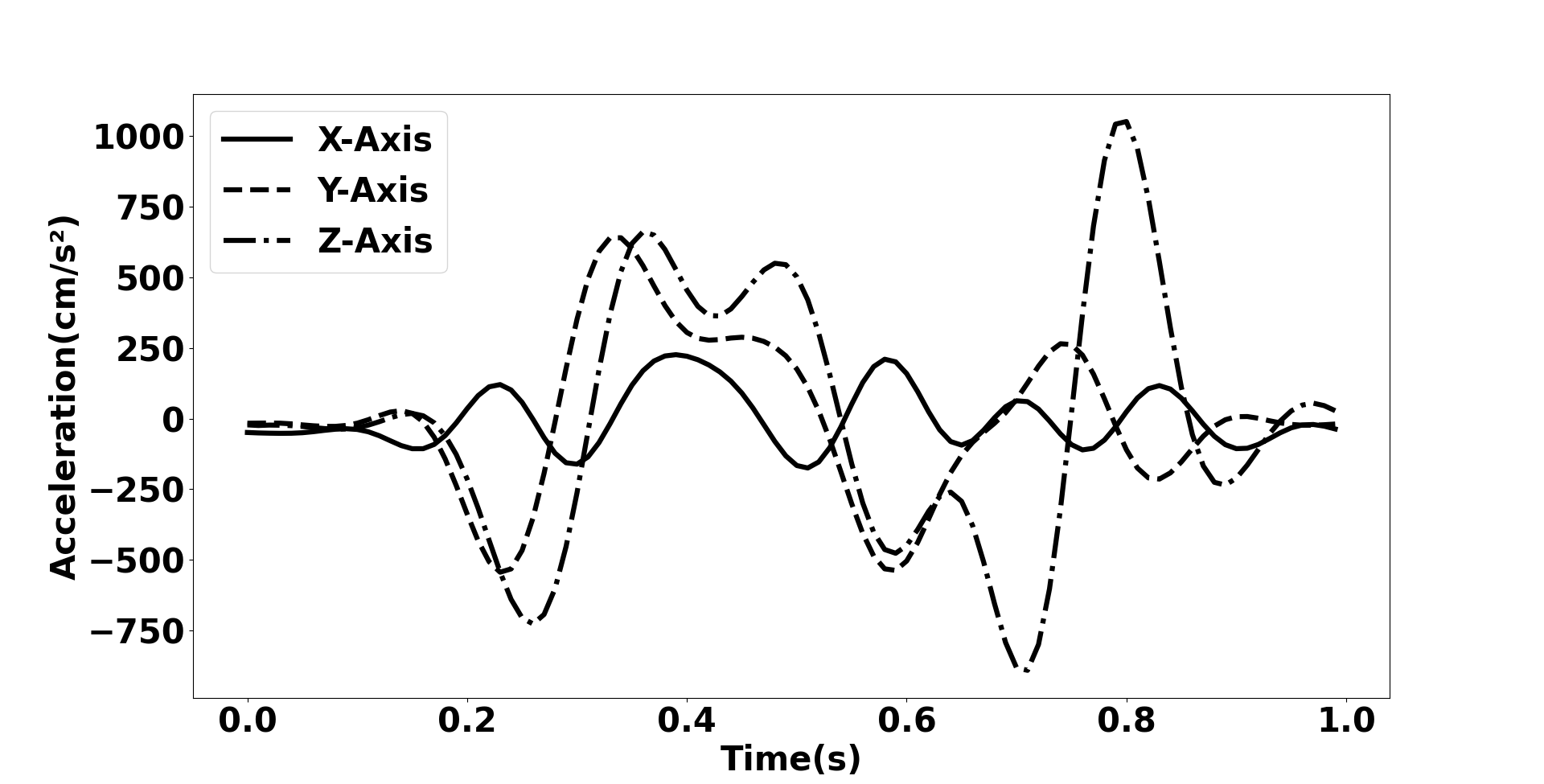}
        \caption{Robot acceleration.}
    \end{subfigure}
    \begin{subfigure}[t]{0.4\textwidth}
        \includegraphics[width=\linewidth]{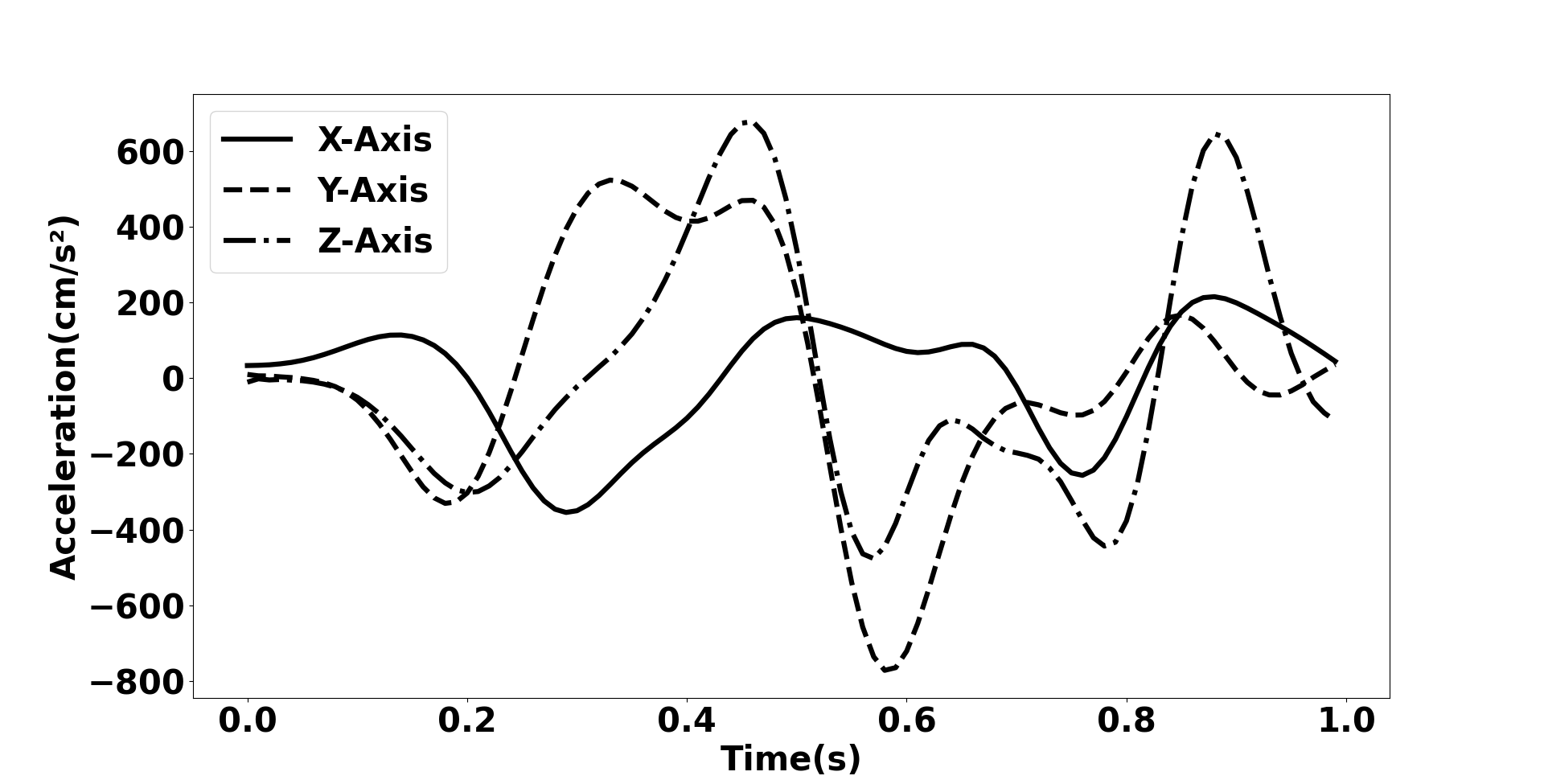}
        \caption{Human acceleration.}
    \end{subfigure}
\bigskip    
    \begin{subfigure}[t]{0.4\textwidth}
        \includegraphics[width=\linewidth]{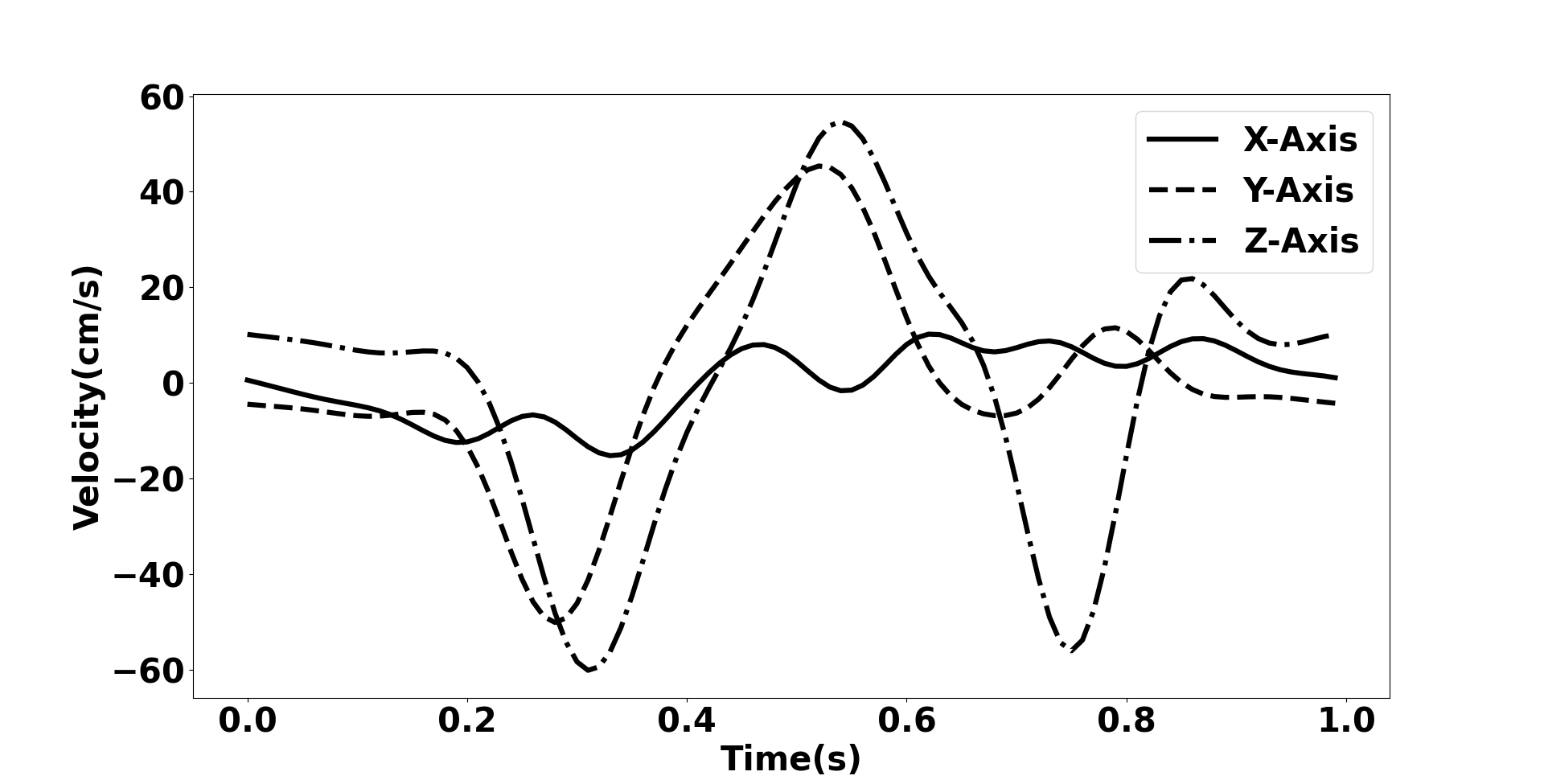}
        \caption{Robot velocity.}
    \end{subfigure}
    \begin{subfigure}[t]{0.4\textwidth}
        \includegraphics[width=\linewidth]{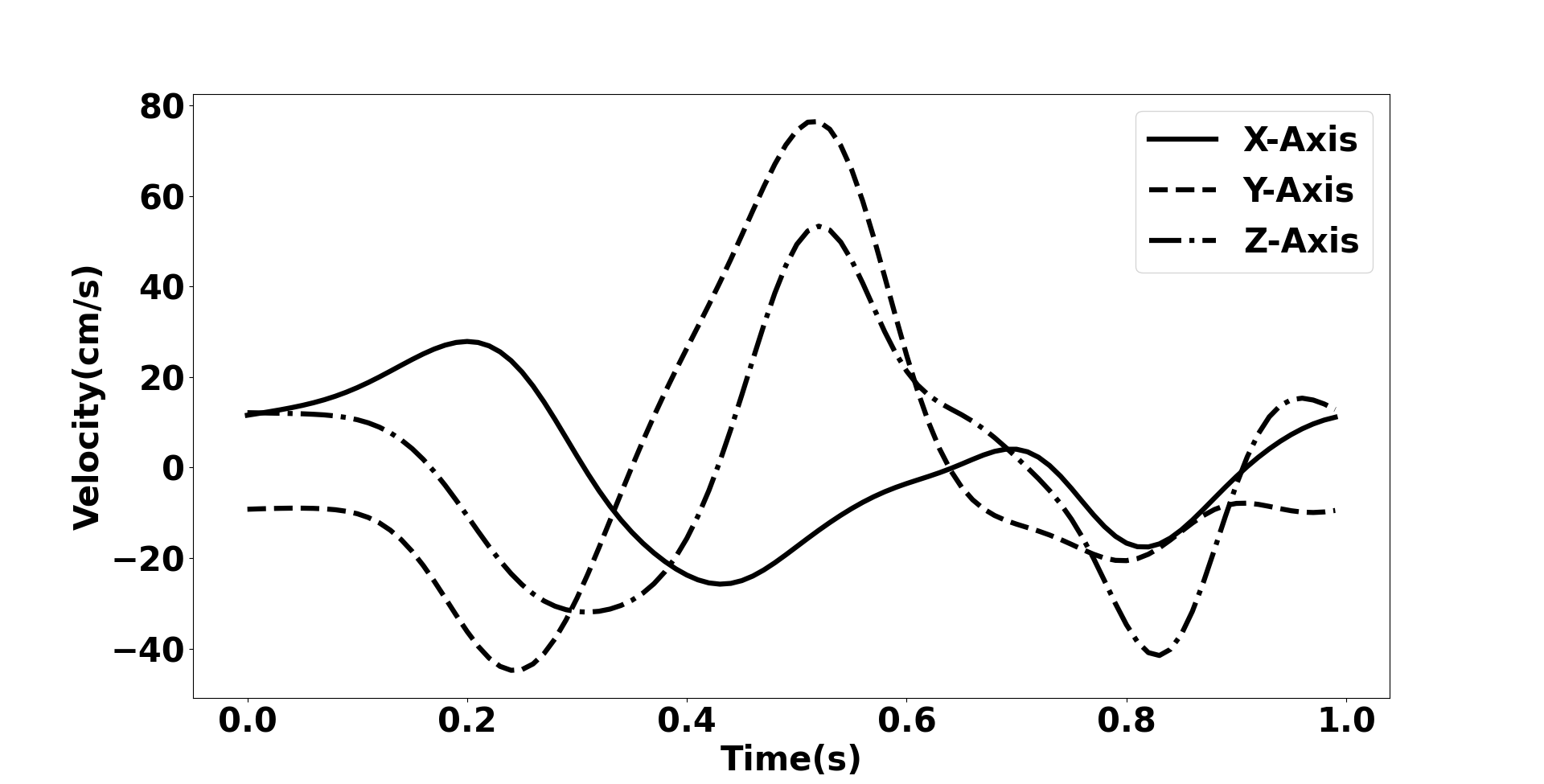}
        \caption{Human velocity.}
    \end{subfigure}
\bigskip   
    \begin{subfigure}[t]{0.4\textwidth}
        \includegraphics[width=\linewidth]{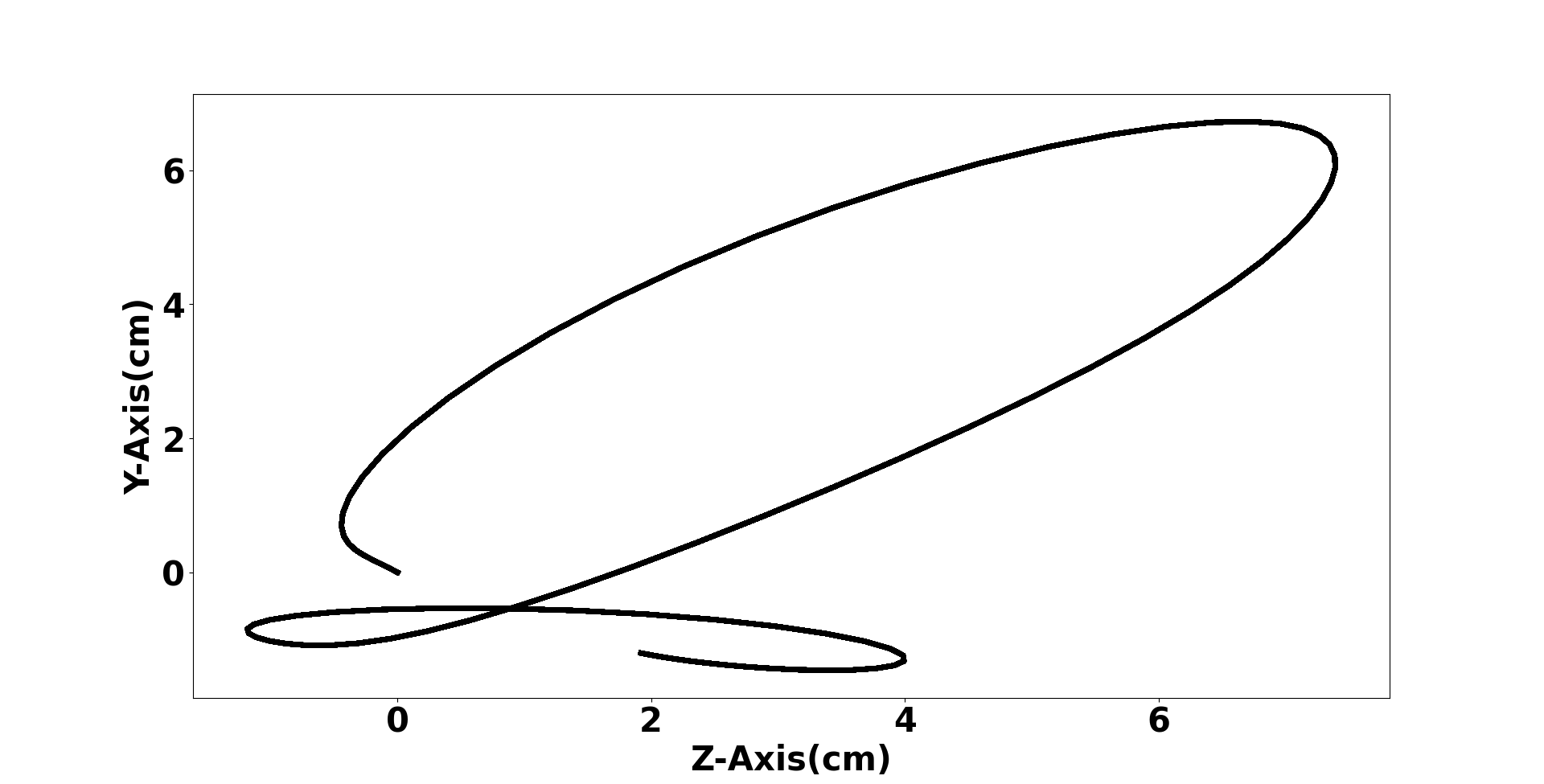}
        \caption{Robot trajectory.}
    \end{subfigure}
    \begin{subfigure}[t]{0.4\textwidth}
        \includegraphics[width=\linewidth]{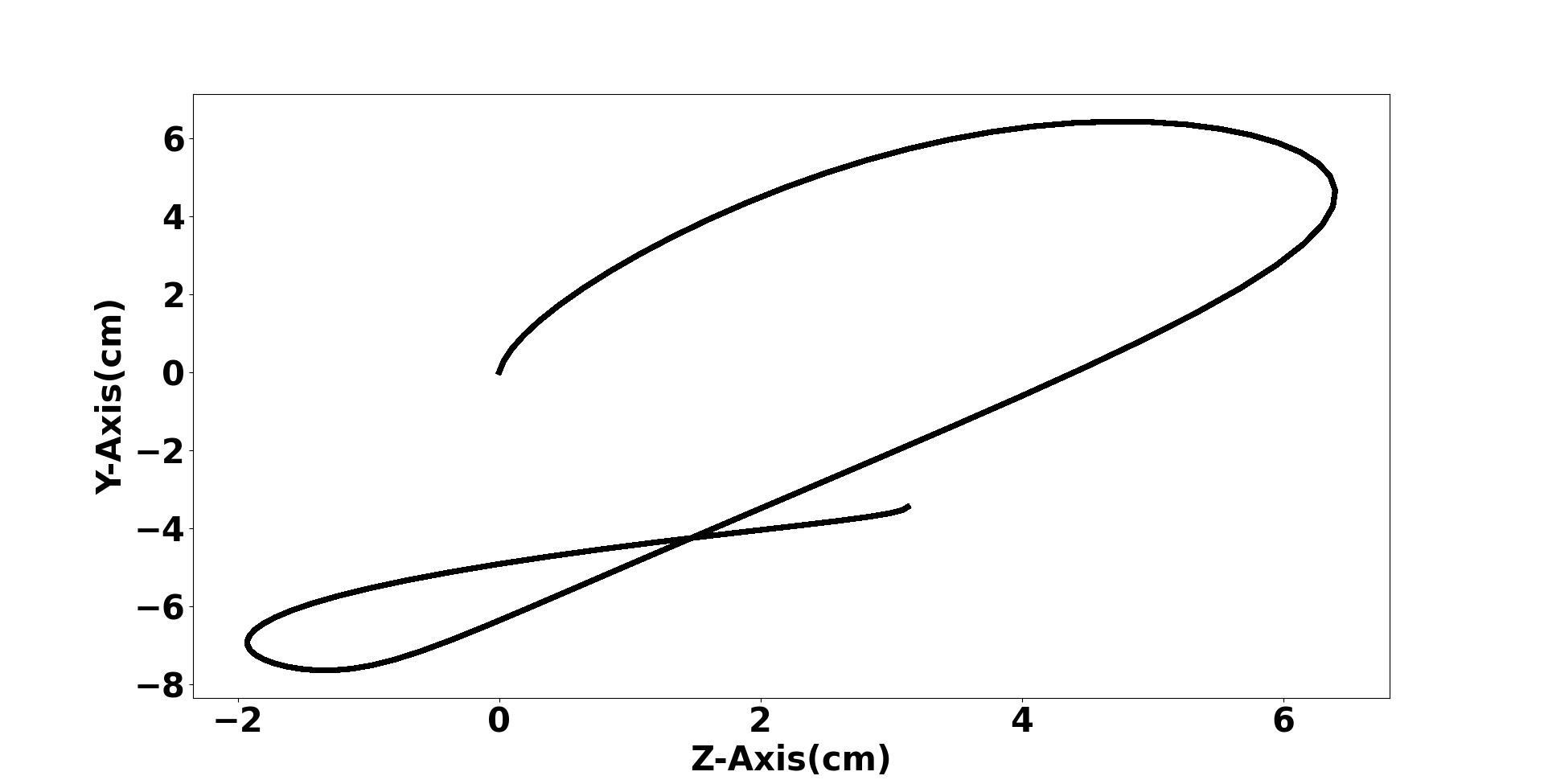}
        \caption{Human trajectory.}
    \end{subfigure}
    \caption{Comparison between Human and Robot's signals}
    \label{fig:allcmp}
\end{figure}

This data was segmented into individual characters and also passed through a low-pass filter of 20 Hz in order to reduce the noise created by the residual movements of both the human and the robot. In Figure \ref{fig:allcmp}, we can observe the signals obtained reproducing the number 2 by a human and by a robot. Each sample had a length ranging from 2 to 4 seconds. 
In order to use a simple classifier, each file was downsampled to 100 samples in each axis using the Fourier method, for a total of 300 samples per file.

\subsection{Classification}

We have chosen to use a multilayer perceptron (MLP) as our classifier, as it is commonly used as a classifier and has been used previously in gesture classification (\cite{Holzbock2023570, Hossain2023, Amit2022292}). The chosen MLP has 3 layers. An input layer, a hidden layer, and an output layer. The nodes on the hidden layer were selected using experimental data to find an optimal distribution. Since we are working with a small volume of data, we have opted against using a larger MLP or other classifiers, since it could lead to overtraining.

The adopted MLP architecture receives 300 inputs. Three different sets of inputs were evaluated. These are the acceleration, the velocity, and the trajectory, all in the three cartesian axes. The first 100 inputs are the X axis, the second 100 are the Y axis, and the last 100, are the Z axis. Three different models were created, all with the same architecture, in order to compare the three signals as inputs.

In order to train the MLP, each model was trained 100 times. Each iteration was an independent training, with 20 epochs, with an early stopping after 10 epochs without improvements, and where a random 20\% of the robotic data was selected for validation, and the 80\% left was used as training data. At each iteration, the initial weights of the network were the ones obtained from the previous training. The human data was only used for the test dataset, in order to avoid possible bias in the training. In each iteration, the accuracy, the loss, and the confusion matrix were obtained. After the training was finished, the average and standard deviation were calculated for accuracy.

\section{Results}

After training the MLP using the acceleration, velocity, and trajectory, confusion matrixes shown in Tables \ref{tab:confaccel}, \ref{tab:confvel} and  \ref{tab:conftraj} were computed. 

If we utilize the acceleration data, as presented in Table \ref{tab:confaccel}, the classifier demonstrates the highest accuracy in correctly classifying characters 0, 4, 1, and 5, all of which have over 69\% of accurate classifications during the training process. However, characters 9 and 6 are the least accurately classified, with less than 1\% accuracy in the case of character 9, as it frequently misclassifies the 9 as a 4.

Using the velocity, as shown in Table \ref{tab:confvel}, it classifies with less than a 10\% error the numbers 5, 2, and 0. In this case, the numbers from 6 to 9 are the less accurate, this time with an average of about 30\% accuracy in each number.

Lastly, with the trajectory in Table \ref{tab:conftraj}, if we compare with the velocity results  we have some improvements in the second half of the numbers, with the exception being the 6, with it being heavily mistaken for the 0. As for the first half, they all have lost some accuracy, with the 3 being the most affected overall.

Comparing the 3 tables, using the velocity for classifying the characters gives us slightly better results than using the acceleration or the trajectory.
 %
\begin{table}[!htbp]
    \centering
    
 \caption{Acceleration}
 
\begin{adjustbox}{width=0.8\linewidth,center}

    \begin{tabular}{| c |c c c c c c c c c c|}
        \hline
        & 0 & 1 & 2 & 3 & 4 & 5 & 6 & 7 & 8 & 9\\
        \hline
         0 & \textbf{90\%} & 0\% & 0\% & 1\% & 9.9\% & 1.2\% & 76\% & 0\% & 0\% & 0\% \\
         1 & 0\% & \textbf{79.7\%} & 0.6\% & 2.3\% & 9\% & 0.4\% & 0\% & 65\% & 0.2\% & 9.6\% \\
         2 & 0\% & 0.6\% & \textbf{42.4\%} & 14.6\% & 0\% & 0\% & 0\% & 2.4\% & 39.3\% & 0.4\% \\
         3 & 9.9\% & 0\% & 11.1\% & \textbf{59.1\%} & 0\% & 15.6\% & 0\% & 0\% & 0\% & 0\% \\
         4 & 0\% & 1.4\% & 0\% & 0\% & \textbf{78.9\%} & 0\% & 0\% & 0.7\% & 12.1\% & 83.6\% \\
         5 & 0\% & 0\% & 0\% & 9.2\% & 0.1\% & \textbf{69.3\%} & 0\% & 0\% & 0\% & 4.9\% \\
         6 & 0.1\% & 0.1\% & 14.8\% & 11.4\% & 0\% & 0\% & \textbf{24\%} & 0\% & 0\% & 0\% \\
         7 & 0\% & 18.2\% & 15.8\% & 0\% & 0.1\% & 0\% & 0\% & \textbf{31.9\%} & 0\% & 0\% \\
         8 & 0\% & 0\% & 10\% & 2.4\% & 1.6\% & 13.5\% & 0\% & 0\% & \textbf{38.5\%} & 0.7\% \\
         9 & 0\% & 0\% & 5.3\% & 0\% & 0.4\% & 0\% & 0\% & 0\% & 9.9\% & \textbf{0.8\%} \\
        \hline
        Total & 100\% & 100\% & 100\% & 100\% & 100\% & 100\% & 100\% & 100\% & 100\% & 100\%\\
        \hline
    \end{tabular}
     \end{adjustbox}
        \centering
    Average accuracy: 51.46\%
    STD: 28.60

    \label{tab:confaccel}
 
\end{table}

\begin{table}[!htbp]
    \centering
       \caption{Velocity}
    \begin{adjustbox}{width=0.8\linewidth,center}
    \begin{tabular}{| c |c c c c c c c c c c|}
        \hline
        & 0 & 1 & 2 & 3 & 4 & 5 & 6 & 7 & 8 & 9\\
        \hline
         0 & \textbf{90\%} & 0\% & 0\% & 0\% & 0.9\% & 0\% & 69.4\% & 0\% & 0.9\% & 0\% \\
         1 & 0\% & \textbf{75\%} & 0\% & 23.5\% & 9.1\% & 0\% & 0\% & 70.9\% & 0\% & 9.5\% \\
         2 & 0\% & 0\% & \textbf{91.1\%} & 0\% & 10\% & 0\% & 0\% & 0\% & 68.6\% & 0\% \\
         3 & 9.7\% & 0\% & 0\% & \textbf{75.6\%} & 0\% & 0\% & 0\% & 0\% & 0\% & 0\% \\
         4 & 0\% & 0\% & 0\% & 0\% & \textbf{80\%} & 0\% & 0.1\% & 0\% & 0\% & 53.8\% \\
         5 & 0\% & 0\% & 0\% & 0\% & 0\% & \textbf{99.5\%} & 0.1\% & 0\% & 0\% & 0.8\% \\
         6 & 0.3\% & 0\% & 0.3\% & 0\% & 0\% & 0\% & \textbf{30.3\%} & 0\% & 0.2\% & 0\% \\
         7 & 0\% & 6.6\% & 8.6\% & 0\% & 0\% & 0\% & 0\% & \textbf{29.1\%} & 0\% & 0\% \\
         8 & 0\% & 0\% & 0\% & 0.6\% & 0\% & 0\% & 0.1\% & 0\% & \textbf{30.3\%} & 0\% \\
         9 & 0\% & 18.4\% & 0\% & 0.3\% & 0\% & 0.5\% & 0\% & 0\% & 0\% & \textbf{35.9\%} \\
        \hline
        Total & 100\% & 100\% & 100\% & 100\% & 100\% & 100\% & 100\% & 100\% & 100\% & 100\%\\
        \hline
    \end{tabular}
        \end{adjustbox}
        
    Average accuracy: 63.68\%
   TD: 28.79
 
    \label{tab:confvel}
\end{table}

\begin{table}[!htbp]
    \centering
      \caption{Trajectory}
    \begin{adjustbox}{width=0.8\linewidth,center}
    \begin{tabular}{| c |c c c c c c c c c c|}
        \hline
        & 0 & 1 & 2 & 3 & 4 & 5 & 6 & 7 & 8 & 9\\
        \hline
         0 & \textbf{85.9\%} & 0\% & 0\% & 0\% & 0\% & 0\% & 81.4\% & 0\% & 27.6\% & 0\% \\
         1 & 0\% & \textbf{71.8\%} & 0\% & 0\% & 25.5\% & 0\% & 0\% & 38.3\% & 0\% & 10\% \\
         2 & 0\% & 0\% & \textbf{80.3\%} & 0\% & 0\% & 0\% & 0\% & 0\% & 0\% & 0\% \\
         3 & 1.6\% & 0\% & 0\% & \textbf{28.8\%} & 3.9\% & 0\% & 0.8\% & 0\% & 0.6\% & 0\% \\
         4 & 0\% & 0\% & 0\% & 0\% & \textbf{52.8\%} & 0\% & 0\% & 0\% & 0\% & 30.7\% \\
         5 & 4.4\% & 8.6\% & 0\% & 0\% & 5.5\% & \textbf{99\%} & 0\% & 5.2\% & 0\% & 7.4\% \\
         6 & 0.2\% & 0\% & 8.1\% & 0\% & 0\% & 0\% & \textbf{8\%} & 0\% & 2.7\% & 0\% \\
         7 & 5.6\% & 10.1\% & 0\% & 45.4\% & 2.1\% & 0\% & 0\% & \textbf{56.5\%} & 0\% & 12.6\% \\
         8 & 2.3\% & 0\% & 11.6\% & 25.8\% & 0.4\% & 1\% & 9.8\% & 0\% & \textbf{69.1\%} & 0\% \\
         9 & 0\% & 9.5\% & 0\% & 0\% & 9.8\% & 0\% & 0\% & 0\% & 0\% & \textbf{39.3\%} \\
        \hline
        Total & 100\% & 100\% & 100\% & 100\% & 100\% & 100\% & 100\% & 100\% & 100\% & 100\%\\
        \hline
    \end{tabular}
        \end{adjustbox}
        
    Average accuracy: 59.15\%
    STD: 27.86
  
    \label{tab:conftraj}
\end{table}

\section{Conclusion}


In this paper, we have explored the feasibility of using robotic movement data as a substitute for training data. Although our results are lower compared to other studies on classifying numeric characters, it is important to note that our study focuses on determining whether it is possible to classify human movement using only robotic motion. With further investigation in this area, we expect to obtain better results, demonstrating the feasibility of using non-human data in human motion databases.

Currently, we are working on improving the results by conducting a more in-depth analysis of human movements and refining the calibration of the robotic arm to accurately replicate the anatomy of the human hand and arm, as humans don't write in a vertical plane with their arm fully extended naturally, as our natural position when we write in a vertical plane is with the elbow bent and a slight rotation and inclination in the wrist. Additionally, we are experimenting with different classification systems, such as Dynamic Time Warping based classifiers or Support Vector Machines, to examine how different classifiers may respond, as well as studying how using different proportions of human and robot data affect the classification systems

\section{Acknowledgement}
This work has been supported by the Spanish project PID2021-122687OA-I00 / AEI  / 10.13039 / 501100011033 /FEDER, UE, and the  Canary Islands Employment Service, Investigo Programme  32/39/2022-0923131539, Recovery, Transformation and Resilience Plan - NextGeneration EU.
%
%
\bibliographystyle{apalike}
\bibliography{ref.bib}
%

\end{document}